\documentclass{bmvc2k}

\usepackage{bm}
\usepackage[dvipsnames]{xcolor}

\title{Single-view Object Shape Reconstruction Using Deep Shape Prior and Silhouette}

\addauthor{Kejie Li}{kejie.li@adelaide.edu.au}{12}
\addauthor{Ravi Garg}{ravi.garg@adelaide.edu.au}{12}
\addauthor{Ming Cai}{ming.cai@adelaide.edu.au}{12}
\addauthor{Ian Reid}{ian.reid@adelaide.edu.au}{12}

\addinstitution{
 The School of Computer Science\\
 University of Adelaide\\
 Adelaide, Australia
}
\addinstitution{
 Australian Centre for Robotic Vision\\
 Adelaide, Australia
}

\runninghead{LI, GARG, CAI, REID}{Single-view Object Shape Reconstruction}

\def\eg{\emph{e.g}\bmvaOneDot}

\def\etal{\emph{et al}\bmvaOneDot}

\begin{document}

\maketitle

\begin{abstract}
3D shape reconstruction from a single image is a highly ill-posed problem. Modern deep learning based systems try to solve this problem by learning an end-to-end mapping from image to shape via a deep network. In this paper, we aim to solve this problem via an online optimization framework inspired by traditional methods. Our framework employs a deep autoencoder to learn a set of latent codes of 3D object shapes, which are fitted by a probabilistic shape prior using Gaussian Mixture Model (GMM). At inference, the shape and pose are jointly optimized guided by both image cues and deep shape prior without relying on an initialization from any trained deep nets. Surprisingly, our method achieves comparable performance to state-of-the-art methods even without training an end-to-end network, which shows a promising step in this direction.
\end{abstract}

\section{Introduction}
Humans effortlessly infer the 3D structure of objects from a single image, including the parts of the object which are not directly visible in the image. It is obvious that the human perception of object geometry involves, in large part, the vast experience of having seen the objects of a category from multiple-views, having reasoned about their geometry and having built rich prior knowledge about meaningful object structure for a specific object category. However, single view object shape reconstruction remains a challenging problem for modern vision algorithms -- perhaps because of the difficulty of capturing and representing this prior shape information -- and the solutions to 3D geometric perception have been traditionally dominated by multi-view reconstruction pipelines. 

The recent successes of deep learning have motivated researchers to depart from traditional multi-view reconstruction methods and directly learn the image to object-shape mapping using CNNs \cite{girdhar2016learning,choy20163d,fan2017point,tulsiani2017multi,tulsiani2018multi,wu2016learning,wu2017marrnet}.
These deep networks show impressive results, with prior information about shape captured at training time and represented implicitly in the network's weights. However, the lack of real images with corresponding ground-truth 3D shapes forces these methods mostly to rely on synthetic images rendered from a object CAD model repository for training, which obviously leads to a domain gap when evaluating on real images.

An alternative way to reason about the 3D geometry of the object is by combining the 2D image cues (i.e.\ object silhouette) with explicitly learned priors about the object shape -- thereby separating the 2D recognition from 3D reconstruction. This two-stage framework has been successfully used for inferring 3D shape from images prior to the resurgence of deep learning by \cite{prisacariu2012simultaneous,prisacariu2011shared} given a single view and \cite{shapeprior,dame2013dense} in a multi-view scenario.
They propose to estimate the 3D object shape and pose given silhouette(s). To make this highly ill-posed problem well-posed, these works learned a probabilistic latent space embedding using a Gaussian Process Latent Variable Model (GPLVM). An image cue based energy function is minimized with respect to the object shape and pose.

These methods worked quite well for object categories with small shape variations and which lack high frequency detail -- like cars or human bodies. Nevertheless, more complex objects like chairs, with larger shape variation and thin structures, were significant failure cases. Some of the limitations arise due to the use of GPLVM for dimensionality reduction. GPLVM does not scale well to be trained on large datasets or to learn priors on high-dimensional shape representations. This forces \cite{prisacariu2012simultaneous} to use frequency based compression techniques for the representing shapes leading to loss of thin and other high frequency object structures. 

Throughout this paper, we aim to answer one important question in the field of single-view object shape reconstruction: Is learning an end-to-end deep network using enormous number of synthetic images really necessary? Although they show promising results in recent years, not only they suffer from domain gap, but also they are expensive to train in terms of synthetic image rendering and network training time. More importantly, one common drawback of deep neural network is the non-transparency property being a black box, which leads to a problem that geometry constraints (i.e.\ silhouette consistency and photometric consistency) that traditional approaches use for geometry reconstruction may not be enforced at inference regardless of whether those constraints are used during training.

To answer the question aforementioned, we present an approach that takes advantage of both deep learning and traditional optimization based methods. Technically, we train a point cloud deep autoencoder that captures a latent space of object shapes, which is further modeled by a Gaussian Mixture Model (GMM). At inference, an object shape is optimized in the form of latent code by jointly maximizing the probability explained by the GMM and minimizing the discrepancy between its projection to image plane using object pose and an object silhouette estimated from an off-the-shelf semantic segmentation network (so-called silhouette consistency). 
Note that unlike single-view 3D reconstruction, a semantic segmentation network can easily be trained on real-image dataset (\eg{MSCOCO} \cite{lin2014microsoft}), thereby domain gap is avoided. 

\begin{figure}[t]
\includegraphics[width=\textwidth]{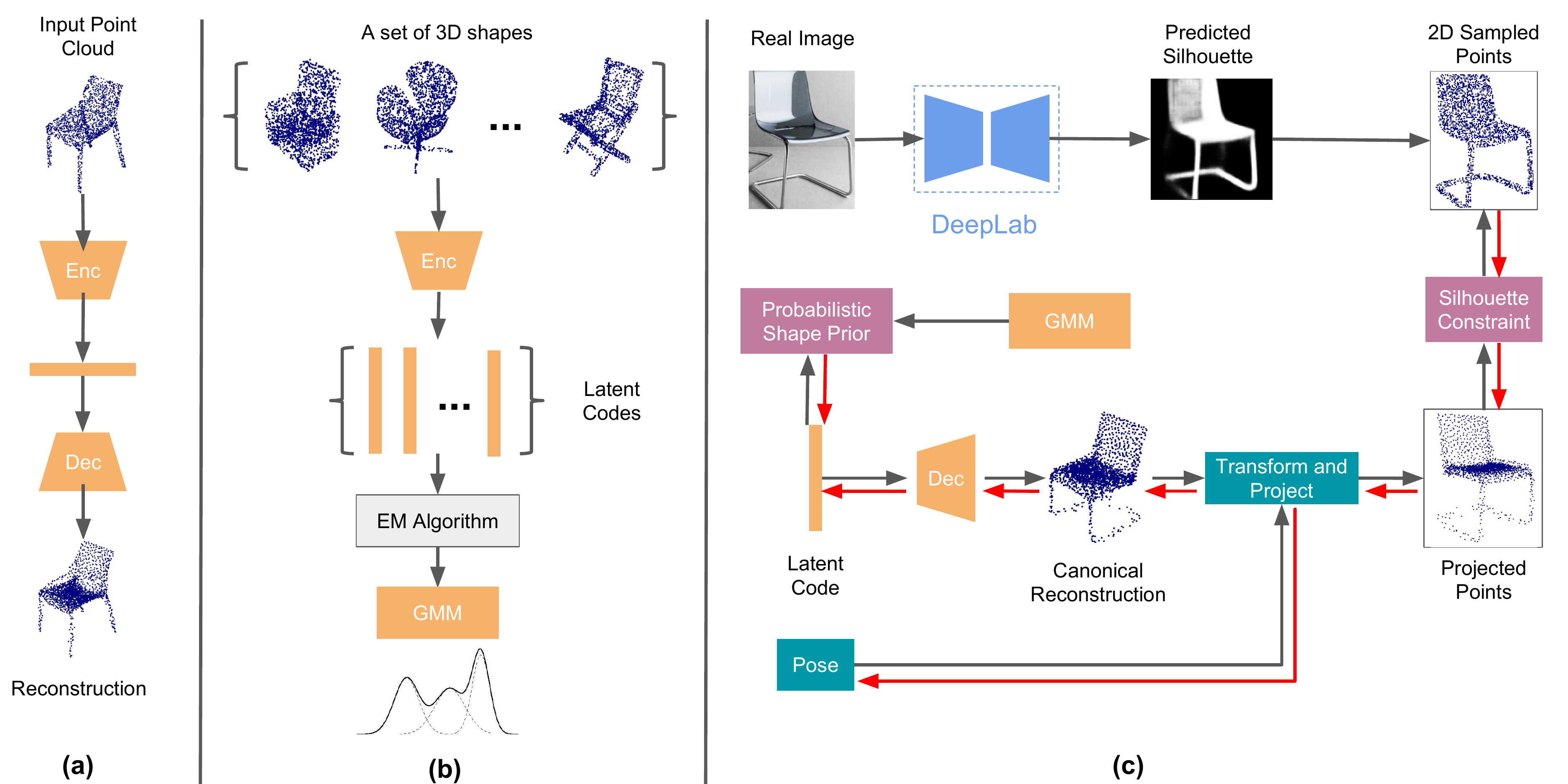}
\caption{(a) We train the point cloud autoencoder to learn a latent space for 3D shapes (b) A set of 3D point clouds are mapped to the latent codes by the trained point cloud encoder. A GMM as a probabilistic model of the latent space is learned by fitting on this set of latent codes. (c) At inference time, we jointly estimate the shape and pose guided by the probabilistic shape prior via the learned GMM and the silhouette-shape constraint.  The \textcolor{Orange}{\textbf{orange}}, \textcolor{SeaGreen}{\textbf{green}}, \textcolor{ProcessBlue}{\textbf{blue}}, and  \textcolor{Fuchsia}{\textbf{purple}} blocks correspond to the components for shape, pose, silhouette, and loss terms respectively. 
The red arrows \textcolor{red}{$\bm{\longrightarrow}$} represent the gradient flow with respect to the latent code and the object pose during inference.}
\label{fig:framework_pipeline}
\end{figure}

Although some approaches \cite{wu2017marrnet,zhu2018object,zhu2017rethinking} also tried to enforce silhouette consistency at inference time, our method is significantly different from them in twofold. Firstly, we initialize the shape optimization using the mean latent codes of the GMM, whereas previous approaches rely on a pre-trained image-to-shape deep network for initialization. Secondly, additional to the silhouette consistency, the search of object shape is constrained by a shape prior term via the GMM, which empirically shows its efficacy in our ablation study. The proposed approach not only achieves comparable performance with the state-of-the-art end-to-end network, but also enjoys the advantage of simplicity and flexibility. An illustration of our framework can be found on Figure \ref{fig:framework_pipeline}. 


In summary, our contributions are as follows:
\begin{itemize} 
    \item By leveraging the power of modern deep learning for non-linear dimensionality reduction, we show that a simple GMM fitting into this latent space can effectively be a probabilistic shape prior that plays a crucial role in our optimization framework.
    \item While pioneering works \cite{wu2017marrnet,zhu2018object} show that online optimization can further improve the one-shot network prediction, to the best of our knowledge, we present the first method that does not rely on a deep network to initialize the shape and pose optimization, which makes our method simpler.
    \item Even without training an end-to-end network, our shape-prior aware inference framework achieve comparable performance to deep learning based prior art for single-view object shape reconstruction on a real image benchmark dataset. 
\end{itemize}

\section{Related Works}
3D understanding of object shapes from images is considered an important step in scene perception. While mapping an environment using Structure from Motion \cite{tomasi1992shape,agarwal2009building} and SLAM \cite{davison2007monoslam,newcombe2011dtam} facilitates localization and navigation, a higher level of understanding about objects in terms of their shape and relative position with respect to the rest of the background is essential for their manipulation. Initial works, to localize objects while estimating its pose from an image, are limited to the case where a pre-scanned object structure is available \cite{rosenhahn2007three,gall2007clustered,schmaltz2007occlusion}.

Soon it is realized that the valid 3D shapes of objects belonging to a specific category are highly correlated and dimensionality reduction emerges as a prominent tool to model object shapes. 
Works like \cite{vicente2014reconstructing} reconstruct traditional image segmentation datasets like PASCAL VOC \cite{everingham2010pascal} by extending the ideas of non-rigid Structure from Motion \cite{bregler2000recovering,brand2001morphable,dai2014simple,garg2013dense}.
Methods like \cite{kar2015category} propose to use these learned category specific object shape manifolds to reconstruct shape of the object from single image by fitting the reconstructed shape to a single image silhouette and refine it with simple image based methods like shape from shading \cite{horn1989obtaining}. 

While the methods mentioned above attempt to learn linear object shape manifolds from 2D key-point correspondences or object segmentation annotations, increasing availability of class specific 3D CAD models and object scans allow researchers to learn complex manifolds for shapes within an object category directly from data. For example, Gaussian Process Latent Variable Model or Kernel/simple Principal Component Analysis is employed to learn a compact latent space of object shapes in \cite{dambreville2008framework,prisacariu2012simultaneous,dame2013dense,engelmann2016joint}. 
Taking advantage of modern deep learning techniques, deep auto-encoders \cite{hinton2006reducing,vincent2008extracting}, VAEs \cite{kingma2013auto}, and GANs \cite{goodfellow2014generative,wu2016learning} can be better alternatives to model object shapes than traditional dimensionality reduction tools. 

Recently, with the advance of deep learning, there are many end-to-end systems proposed to solve the object shape reconstruction problem from a single RGB image directly. Different supervision has been explored. Direct 3D supervision is dominant in the early works \cite{choy20163d,girdhar2016learning,fan2017point,wu2017marrnet,groueix2018atlasnet}. Weaker supervision that uses geometry projection (i.e.\ depth consistency, silhouette consistency) is later employed in \cite{tulsiani2017multi,tulsiani2018multi,insafutdinov18pointclouds,lin2018learning,li2018efficient,zhu2017rethinking,wiles2017silnet,henderson2018learning,yan2016perspective} as one step moving toward training the network using real images. However, all multiple views of an object are required to train these networks, and they still rely on a one-shot prediction from the network at inference time.

Most related to our work are CodeSLAM \cite{bloesch2018codeslam}, MarrNet \cite{wu2017marrnet}, and Zhu \etal \cite{zhu2018object}, which also use a low-dimensional latent space learnt by a deep network to represent the geometry information. For the scene-level geometry reconstruction, CodeSLAM uses nearby views to form a photometric loss, which is minimized by searching in the learned latent space of depth maps. MarrNet \cite{wu20183d} and Zhu \etal \cite{zhu2018object} enforce the 2.5D (silhouette, depth or normal) - 3D shape constraint and photometric loss respectively to search in the latent space of object shapes. Our work differs from theirs in twofold. Firstly, a common disadvantage of these works, which we address in our work, is that they do not consider a probabilistic prior in the latent space explicitly when they perform optimization, which we show is essential for the optimization at inference time through our ablation study. Secondly, we do not need to train a deep network to initialize optimization variables (i.e.\ shape latent code and pose)

\section{Method}\label{sec:method}
The goal of this work is to reconstruct a 3D shape of an object from a single image.
Our framework closely follows traditional object prior based 3D reconstruction methods by framing the reconstruction as an optimization problem assuming that a deep neural network provides us with the object silhouette. 
This inference method is described in the Section \ref{sec:optimization}. To facilitate the inference, we require a low-dimensional latent space which encodes the object shape, a learned probabilistic prior on this latent space that estimates the confidence of a latent code. How we learn the relevant prerequisites for inference is elaborated in Section \ref{sec:learning}, followed by implementation details in Section \ref{sec:implement}.

\subsection{Inference-as-optimization}\label{sec:optimization}
For inference, we are given with 
sampled points from the predicted silhouette $\bm{S_{sil}}$, initial latent code of the object shape $\bm{l_0}$, initial pose $\bm{R_0}$. Additionally, the learned probabilistic prior on the latent space is available in the form of a GMM.

We minimize the objective function defined below with respect to the shape and pose variables $\bm{l}$ and $\bm{R}$:
\begin{equation}
    L(\bm{l},\bm{R}) = L_{sil} + \lambda L_{shape}.
\label{eq:objective}
\end{equation}
Here $L_{sil}$ corresponds to the silhouette-shape constraint and $L_{shape}$ is the term related to the probabilistic shape prior, which we explain in the rest of this section. $\lambda$ is the weighting factor between these two loss terms.

\textbf{Silhouette-shape Constraint Term}.
It enforces the projection of a transformed 3D shape to be consistent with the object's silhouette on the image.
While using point cloud representation, this can be achieved by defining a 2D Chamfer loss between projection of the transformed 3D point cloud and sampled points of the predicted silhouette shown in Equation \eqref{eq:sil_objective},
\begin{equation}
\begin{split}
\bm{S} &= dec(\bm{l}), \\
L_{sil}(\bm{P},\bm{S},\bm{S_{sil}}) &= \sum_{\bm{s_i} \in \bm{S}} \min_{\bm{s_j} \in \bm{S_{sil}}} \|\bm{P} \bm{s_i} - \bm{s_j}\|^2+\sum_{\bm{s_j} \in \bm{S_{sil}}} \min_{\bm{s_i} \in \bm{S}} \|\bm{P} \bm{s_i}-\bm{s_j} \|^2,
\end{split}
\label{eq:sil_objective}
\end{equation}
where $\bm{S}$ is a point cloud reconstruction in a predefined canonical pose generated by a latent code $\bm{l}$ through the point cloud decoder. $\bm{P}$ is a transformation matrix that is composed of $\bm{K}[\bm{R}]$, projection matrix $\bm{K}$, object rotation $\bm{R}$ (we use orthogonal projection because camera intrinsic is assumed unknown).
$\bm{S_{sil}}$ is a set of 2D sampled points from a predicted silhouette.

\textbf{Shape Prior Term.}
Because of the highly unconstrained nature of the silhouette-shape constraint and incorrect silhouette, we employ a shape prior term, which quantifies the likelihood of a latent code using a probabilistic model. This can mitigates the problem of generating unrealistic 3D shapes.
Specifically, the shape prior minimizes the negative log likelihood of a latent code in the GMM, which can be written as:
\begin{equation}
    L_{shape}(\bm{l}) = -log(\sum_{k=0}^{K}\pi_k \mathcal{N}(\bm{l}|\bm{\mu_k},\bm{\Sigma_k})),
\label{eq:shape_constraint}
\end{equation}
where $\pi_k$, $\bm{\mu_k}$, and $\bm{\Sigma_k}$ are the weight, mean, and covariance matrix of the $k^{th}$ Gaussian component, and $K$ is the number of Gaussian components.

\textbf{Initialization}
A sensible initialization of the shape latent code is using one of the means of GMM. However, we observed that the optimized shape is susceptible to local minimums if we initialize the shape latent code from one position only due to the non-convexity of the optimization. We therefore run the optimization multiple times with different means of the GMM to alleviate this problem. We initialize the rotation at $0$ degree for azimuth, elevation, and tilt. 

\subsection{Learning}\label{sec:learning}
Our inference method relies on learning a low-dimensional latent space of shape from which a decoder network can map a latent code $\bm{l}$ to a unique 3D point cloud $\bm{S}$. To learn this mapping, we train an autoencoder. Similar to a conventional autoencoder, our encoder $enc(.)$ maps a point cloud $\bm{S}$ to its corresponding low dimensional latent code $\bm{l}$, followed by the decoder $dec(.)$ reconstructs the point cloud $\bm{\hat{S}}$ from the latent code $\bm{l}$. We train this autoencoder using 3D Chamfer Distance as defined below:
\begin{equation}
L_{cd}(\bm{\hat{S}},\bm{S}) = \sum_{\hat{\bm{s_i}} \in \bm{\hat{S}}} \min_{\bm{s_j} \in \bm{S}} \|\hat{\bm{s_i}}-\bm{s_j}\|^2
+\sum_{\bm{s_i} \in \bm{S}} \min_{\hat{\bm{s_j}} \in \bm{\hat{S}}} \|\hat{\bm{s_j}}-\bm{s_i}\|^2 .
\label{eq:chamfer}
\end{equation}

To learn a probabilistic model as the shape prior, we fit a GMM to the learned latent space.
Specifically, we map the point clouds $\{ \bm{S}_1 ... \bm{S}_N \}$ to the latent space using the point cloud encoder, to get a set of latent codes $\{ \bm{l}_1 ... \bm{l}_N \}$. 
The parameters of GMM with $K$ components are then learned to maximize the likelihood of the latent codes by using the EM algorithm to optimize the following objective:

\begin{equation}
    \min_{\substack{\pi_1 ... \pi_K, \\ \bm{\mu_1} ... \bm{\mu_K},\\
    \bm{\Sigma_1} ... \bm{\Sigma_K}}} -\sum_{i=0}^{N} log (\sum_{k=0}^{K} \pi_{k} \mathcal{N}(\bm{l_i} | \bm{\mu_k}, \bm{\Sigma_k})) ,
 \label{eq:gmm_mle}
\end{equation}
where $\bm{\pi_k}$, $\bm{\mu_k}$, and $\bm{\Sigma_k}$ are the weight, mean and covariance matrix of the $k^{th}$ Gaussian component, and $N$ and $K$ are the number of sample latent codes and the number of Gaussian component. 

There are a number of generative models that we considered besides GMM during development. For instance, we experimented with VAE and GAN.
However, we found that VAE fails to reconstruct accurately, which we suspect is due to two reasons. Firstly, one well-known issue of VAE is that the KL divergence is too restrictive \cite{chen2016variational}, which makes the reconstruction uninformative to the input. Secondly, a simple isotropic gaussian prior may not be expressive for complicated and diverse shapes. Although the latent space learnt by GAN can produce realistic shapes, a problem we observe is mode collapse in local region of the latent space, which means that neighboring latent codes correspond to identical shape. It makes traversing in the latent space difficult for our gradient based optimization. Other drawbacks of using GAN are also revealed by \cite{achlioptas2017latent_pc}, which empirically show that, in the form of point cloud, among variants of GAN, and GMM, the latter outperforms other models in generalization and diversity of shapes, meaning that the GMM captures a better distribution of object shapes.

\subsection{Implementation Details}\label{sec:implement}
The architecture of our encoder is similar to PointNet \cite{qi2017pointnet}. We use five 1D convolution layers, followed by a max pooling as a ``permutation invariance'' function, and two fully connected layers, to map from a point cloud (a $2048\times3$ matrix) to a latent code. The decoder is composed of five fully connected layers to map from a latent code to a point cloud reconstruction. ReLU \cite{nair2010rectified} is used between layers except the last one as a non-linear activation function. Point clouds used for training are samples from the CAD models of the ShapeNet dataset \cite{chang2015shapenet}. The autoencoder is trained on 13 object classes including chair, table and sofa, and is trained for 150 epochs using Adam optimizer \cite{kingma2014adam}. We use GMM with 5 Gaussian components for chair class and 4 Gaussian components for table and sofa classs when evaluating on Pix3D \cite{sun2018pix3d}. The number of Gaussian components is chosen such that each mean represents a distinctive object shape. The GMM is optimized using the EM algorithm implemented in Python Scipy library \cite{scipy}.

At inference, the object silhouette is given by DeepLab v3 \cite{chen2017rethinking}. Rotation parameters and shape latent code are optimized jointly reusing Adam optimizer in Tensorflow \cite{tensorflow2015-whitepaper}. $\lambda$ during the first 500 iterations is set to a relatively high value at $1$ to reduce the shape-pose ambiguity, and later reduced to $0.001$. The optimization is terminated when the rate of change in loss function diminishes or the maximum number of 1000 iterations is reached. 

\begin{table}
\begin{center}
\begin{tabular}{l|c c}
\hline
Methods & CD $\downarrow$ & EMD $\downarrow$ \\
\hline\hline
3D-R2N2    & $0.239$ & $0.211$ \\
PSGN       & $0.200$ & $0.216$ \\
3D-VAE-GAN & $0.182$ & $0.176$ \\
DRC        & $0.160$ & $0.144$ \\
MarrNet    & $0.144$ & $0.136$ \\
AtlasNet   & $0.125$ & $0.128$ \\
\textbf{Ours}  & \bm{$0.116$} & $0.125$ \\
Pix3D      & $0.119$ & \bm{$0.118$} \\
\hline
\end{tabular}
\end{center}
\caption{3D shape reconstruction results on Pix3D dataset}
\label{tab:pix3d_shape}
\end{table}

\section{Experiments}
We provide experimental results of our framework on single view object shape reconstruction on a real dataset and compare it against prior art. 
A series of ablation studies are also presented.

\subsection{Testing Data and Evaluation Protocol}
We evaluate on a recently published Pix3D \cite{sun2018pix3d} dataset -- a dataset consisting diverse shapes aligned with corresponding real images. We benchmark with prior art on chair class following Pix3D setup, where 2894 real images and 221 corresponding 3D ground-truth shapes are used. We also showcase that our method can adapt to other object classes by experimenting on table and sofa class in the ablation study section.
Other alternatives to Pix3D which were used popularly were either purely synthetic datasets like ShapeNet \cite{chang2015shapenet}, or PASCAL 3D+ \cite{xiang2014beyond} which was created by associating a small number of overlapping 3D CAD models per category for both training and test images of a particular object category to create pseudo ground truth. Due to this reason, as discussed in \cite{tulsiani2017multi,wu2017marrnet} this dataset is not deemed fit for shape reconstruction evaluation recently.

We use 3D Chamfer Distance (CD as defined in \eqref{eq:chamfer}) and Earth Mover Distance (EMD as defined in \eqref{eq:emd}) as evaluation measures for shape reconstruction following the human study on the evaluation metrics for shape reconstruction presented in Pix3D which shows that these matrices are more correlated with human perception than IoU.
\begin{equation}
    EMD(S_1,S_2) = \frac{1}{\|S_1\|} \min_{\phi:S_1 \rightarrow S_2} \sum_{x \in S_1} \|x-\phi(x)\|_2,
\label{eq:emd}
\end{equation}
where $\phi:S_1 \rightarrow S_2$ is a bijection. We use the approximation implemented in Pix3D benchmark setup to calculate the EMD efficiently.

\subsection{Shape Reconstruction on Pix3D}
\begin{figure}[t]
\includegraphics[width=\textwidth]{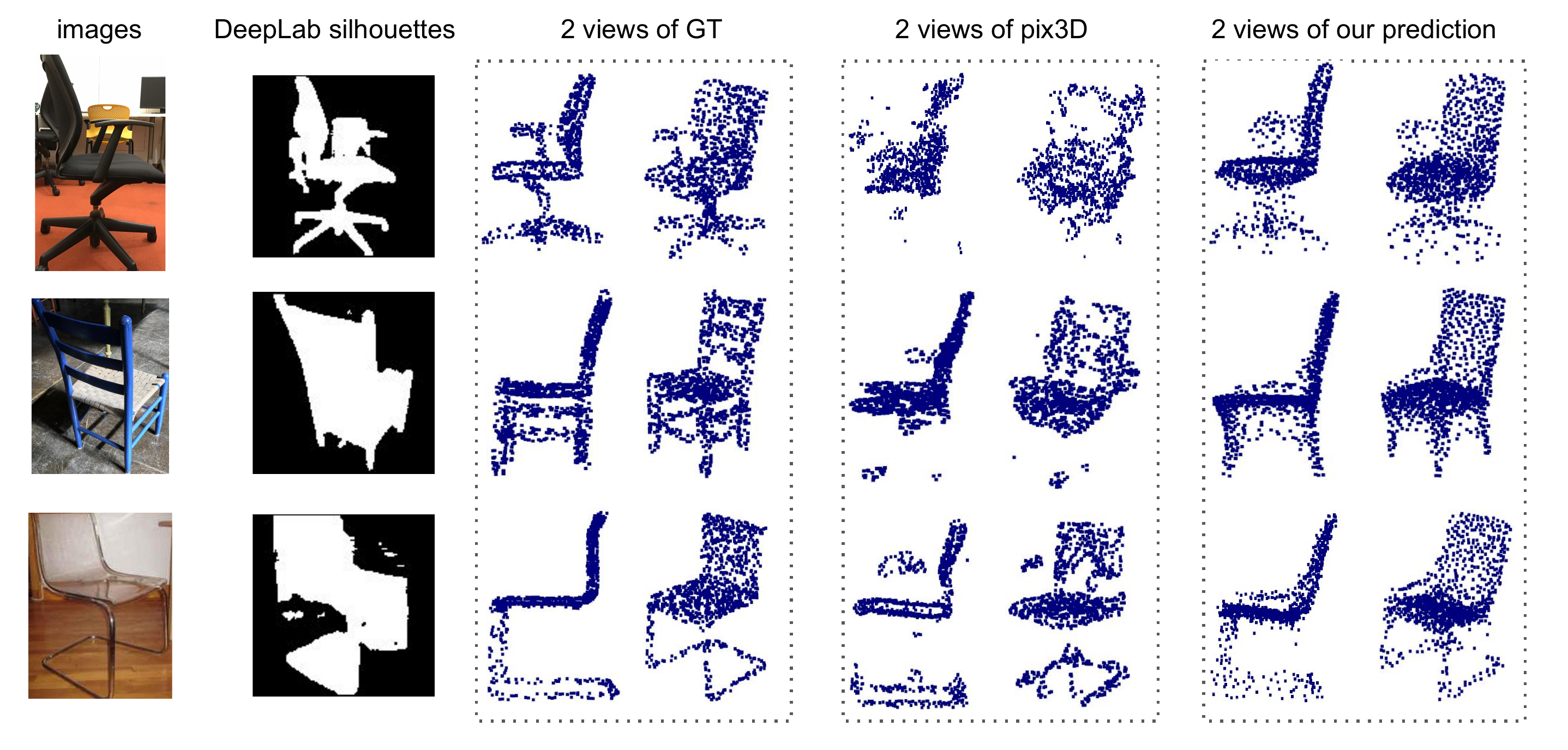}
\caption{Qualitative comparison with Pix3D.}
\label{fig:sota_comparison}
\end{figure}
We compare the performance of our 3D shape prediction pipeline with a list of prior art including 3D-VAE-GAN \cite{wu2016learning}, 3D-R2N2 \cite{choy20163d}, PSGN \cite{fan2017point}, DRC \cite{tulsiani2017multi}, AtlasNet \cite{groueix2018atlasnet}, MarrNet \cite{wu2017marrnet} and Pix3D \cite{wu20183d} which is a variant of MarrNet that decouples shape and pose prediction. The quantitative result is reported in Table \ref{tab:pix3d_shape}. The performance evaluation for most of these methods listed above is taken from Pix3D paper \cite{sun2018pix3d}. Despite the simplicity of our method, which only relies on deep shape prior and silhouette consistency within an optimization framework, we outperform many end-to-end deep networks that are specifically designed to reconstruct object shape from a single view. Qualitative results are shown in Figure \ref{fig:sota_comparison} for visual comparison. We hope this result can facilitate the community to re-think the role of deep learning in the task of single-view object shape reconstruction or even other geometry reconstruction tasks. Is an end-to-end network really necessary or the best practice for single-view shape reconstruction?

\subsection{Ablation Study on Shape prior}\label{sec:ablation}
In this section, we demonstrate that the proposed probabilistic shape prior plays a crucial part in our optimization pipeline. 
To that end, we compare the optimization using silhouette only and our full loss term which takes into account the likelihood of generated shape by using both $L_{sil}$ and $L_{shape}$. 
\begin{table}
\begin{center}
\begin{tabular}{l|c c|c c}
\hline
Loss & \multicolumn{2}{c}{$L_{sil}$ only} & \multicolumn{2}{c}{$L_{sil}$ $+$ $L_{shape}$} \\
\hline
 &CD & EMD & CD & EMD \\
\hline\hline
chair   & $0.137$ & $0.148$ & $\bm{0.116}$ & $\bm{0.125}$ \\
table   & $0.173$ & $0.176$ & $\bm{0.158}$ & $\bm{0.172}$ \\
sofa    & $0.116$ & $0.108$ & $\bm{0.099}$ & $\bm{0.092}$ \\
\hline \hline
\end{tabular}
\end{center}
\caption{Quantitative analysis on the effect of silhouette-shape constraint and probabilistic shape prior at inference.}
\label{tab:abla_shape}
\vspace{-0.3cm}
\end{table}
\begin{figure}[t]
\includegraphics[width=\textwidth]{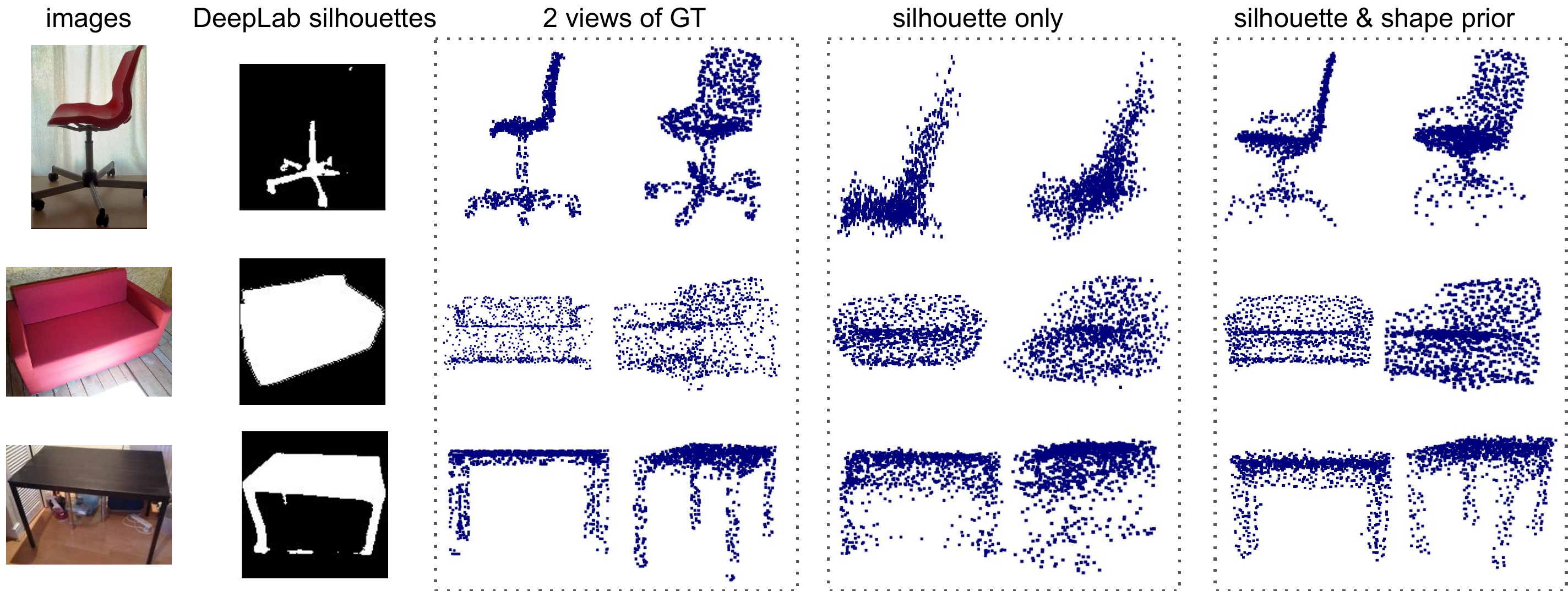}
\caption{Qualitative results using GMM shape prior.}
\label{fig:prior_comparison}
\end{figure}

As for the quantitative comparison in Table \ref{tab:abla_shape}, when we minimize the objective function that penalizes a latent code with low likelihood, the reconstruction performance improves on both measures significantly. 

The main takeaway from these results is that, jointly optimizing shape and pose using silhouette only is a highly ill-posed problem and leads to unrealistic shape with a wrong pose in order to fit the silhouette. The GMM shape prior avoid the regions of the latent space to restrict the inferred shape to be limited to realistic object-like solutions -- thereby avoiding the overfitting to a noisy silhouettes. We present Figure \ref{fig:prior_comparison} for visual comparison.

\begin{figure}[t]
\includegraphics[width=\textwidth]{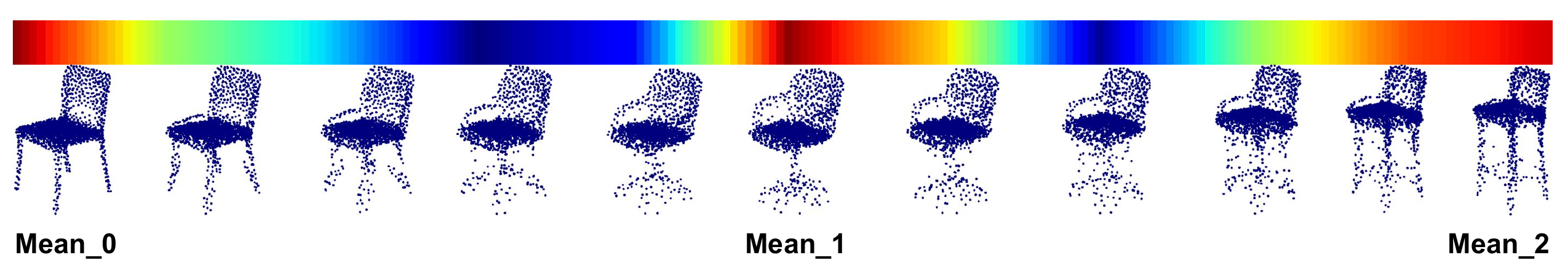}
\caption{This figure visualizes that a ``chair-like'' shape has higher confident in the GMM. We denote the likelihood of a shape explained by the GMM using colors.  Red means high likelihood whereas blue represents low likelihood.}
\label{fig:gmm_interpolate}
\vspace{-0.5cm}
\end{figure}

To further explore this hypothesis, we choose three points from the latent space corresponding to the mean of different Gaussian components and interpolate linearly in the latent space to generate intermediate sample chair shapes. The results are visualized in Figure \ref{fig:gmm_interpolate}. It can be clearly seen that the latent space learned using deep autoencoder training constitutes of regions corresponding to very not chair-like shapes. However, these not chair-like shapes usually correspond to the high variance region of the learned GMM -- for instance, between the latent code of a four-legged chair (mean\_0) and that of a swivel chair (mean\_1), the likelihood of an unrealistic shape as a mix of two structures is particularly low. 

\vspace{-0.2cm}
\section{Discussion}
Instead of training an end-to-end network to address single-view object shape reconstruction, we present an online optimization framework that optimizes the object shape in the form of deep latent code at inference without relying on any deep learning initialization. The latent search space is constrained by a shape prior via a GMM fitted in the latent space and object silhouette given by an off-the-shelf segmentation network, which can be trained with real images and does not suffer from domain gap. This approach achieves comparable state-of-the-art performance on a large dataset of real images when enjoying its simplicity. While this result shows an interesting direction to tackle single-view object shape reconstruction besides training an end-to-end network, there are a number of directions that are worth investigation. For example, instead of fitting a GMM after a latent space is learnt, how to simultaneously learn a latent space and fit a multi-cluster probabilistic shape prior? It would also be interesting to apply our shape-prior aware optimization framework to a multi-view setup for reconstruction and tracking.

\bibliography{reference}
\end{document}